\documentclass[conference]{IEEEtran}

\usepackage[utf8]{inputenc} 

\usepackage[sort,compress]{cite}
\usepackage[pdftex]{graphicx}
\usepackage{amsmath,amssymb}
\usepackage{array}
\usepackage{url}
\usepackage{color,xspace}
\usepackage{adjustbox}
\usepackage{multirow}
\usepackage{tabularx}
\usepackage{booktabs}
\usepackage{listings}
\usepackage[keeplastbox]{flushend}
\usepackage{xfrac}
\usepackage{makecell}

\usepackage{color}
\usepackage{xspace}
\usepackage{hyperref}
\usepackage{algorithm}
\usepackage{algorithmicx}
\usepackage{algpseudocode}

\usepackage{balance}

\newcommand{\vct}[1]{\ensuremath{\boldsymbol{#1}}} 

\newcommand{\set}[1]{\ensuremath{\mathcal{#1}}}

\newcommand{\argmax}{\operatornamewithlimits{\arg\,\max}}

\newcommand{\argmin}{\operatornamewithlimits{\arg\,\min}}

\newcommand{\ie}{{i.e.}\xspace}
\newcommand{\eg}{{e.g.}\xspace}

\newcommand{\etal}{et.al.}
\newcommand{\myparagraph}[1]{\smallskip \noindent \textbf{#1}}

\begin{document}



\title{Deep Neural Rejection against Adversarial Examples}

\author{
	\IEEEauthorblockN{
		Angelo Sotgiu\IEEEauthorrefmark{1},
		Ambra Demontis\IEEEauthorrefmark{1},
		Marco Melis\IEEEauthorrefmark{1},
		Battista Biggio\IEEEauthorrefmark{1}\IEEEauthorrefmark{2},
		Giorgio Fumera\IEEEauthorrefmark{1},				
		Xiaoyi Feng\IEEEauthorrefmark{3},
		and
		Fabio Roli\IEEEauthorrefmark{1}\IEEEauthorrefmark{2}}
	
	\IEEEauthorrefmark{1}University of Cagliari, Italy\\ 
	\{angelo.sotgiu, ambra.demontis, marco.melis, battista.biggio, fumera, roli\}@unica.it \\
	fengxiao@nwpu.edu.cn \\
	\IEEEauthorrefmark{2}Pluribus One, Italy \\
	\IEEEauthorrefmark{3}Northwestern Polytechnical University, Xi'an, China
}

\maketitle

\begin{abstract} 
Despite the impressive performances reported by deep neural networks in different application domains, they remain largely vulnerable to adversarial examples, i.e., input samples that are carefully perturbed to cause misclassification at test time.
In this work, we propose a deep neural rejection mechanism to detect adversarial examples, based on the idea of rejecting samples that exhibit anomalous feature representations at different network layers.
With respect to competing approaches, our method does not require generating adversarial examples at training time, and it is less computationally demanding.
To properly evaluate our method, we define an adaptive white-box attack that is aware of the defense mechanism and aims to bypass it. Under this worst-case setting, we empirically show that our approach outperforms previously-proposed methods that detect adversarial examples by only analyzing the feature representation provided by the output network layer.
\end{abstract}





\begin{figure*}[t]	
	\centering
	\includegraphics[width=.75\textwidth]{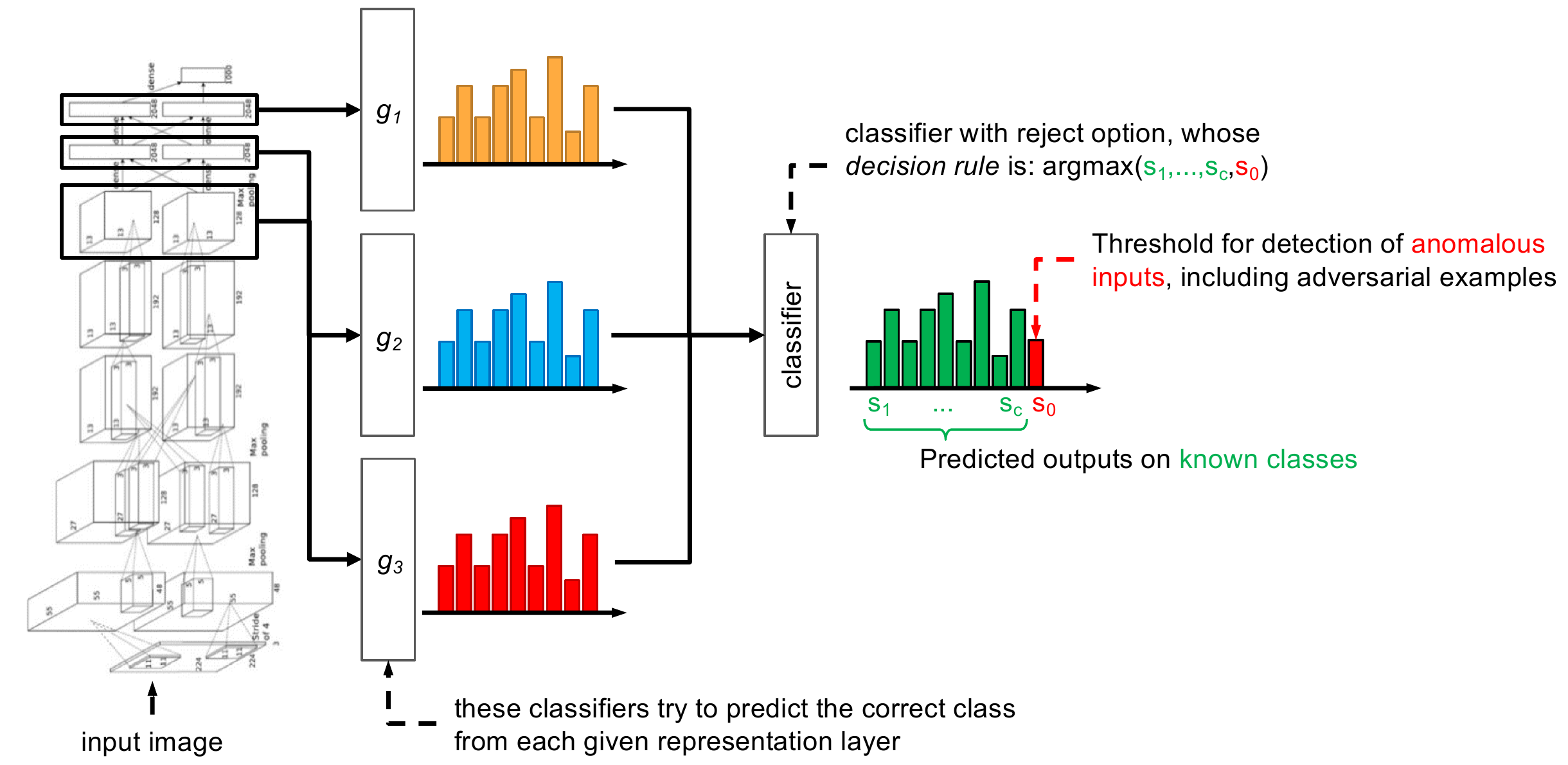}
	\caption{Architecture of Deep Neural Rejection (DNR). DNR considers different network layers and learns an SVM with the RBF kernel on each of their representations. The otputs of these SVMs are then combined using another RBF SVM, which will provide prediction scores $s_1, \ldots, s_c$ for each class. This classifier will reject samples if the maximum score $\max_{k=1, \ldots, c} s_k$ is not higher than the rejection threshold $\theta$. This decision rule can be equivalently represented as $\argmax_{k=0,\ldots,c} s_k(\vct x)$, if we consider rejection as an additional class with $s_0 = \theta$.}
	\label{fig:method}
\end{figure*}

\section{Introduction}

Despite their impressive performances on a variety of tasks, it has been known for more than a decade that machine-learning algorithms can be misled by different adversarial attacks, staged either at training or at test time~\cite{joseph18-advml-book,biggio18}.
After the first attacks proposed against linear classifiers in 2004~\cite{dalvi04,lowd05-ceas}, 
Biggio~\etal~\cite{biggio12-icml,biggio13-ecml} have been the first to show that nonlinear machine-learning algorithms, including support vector machines (SVMs) and neural networks, can be misled by gradient-based optimization attacks~\cite{joseph18-advml-book}.
Nevertheless, such vulnerabilities of learning algorithms have become extremely popular only after that Szegedy~\etal~\cite{szegedy14-iclr,goodfellow15-iclr} have demonstrated that also deep learning algorithms exhibiting superhuman performances on image classification tasks suffer from the same problems.
They have shown that even only slightly-manipulating the pixels of an input image can be sufficient to induce deep neural networks to misclassify its content. Such attacks have then been popularized under the name of \emph{adversarial examples}~\cite{szegedy14-iclr,biggio13-ecml,biggio18}.

Since the seminal work by Szegedy~\etal~\cite{szegedy14-iclr,goodfellow15-iclr}, many defense methods have been proposed to mitigate the threat of adversarial examples. Most of the proposed defenses have been shown to be ineffective against more sophisticated attacks (\ie, attacks that are aware of the defense mechanism), leaving the problem of defending neural networks against adversarial examples still open. According to~\cite{biggio18}, the most promising defenses can be broadly categorized into two families. The first includes approaches based on robust optimization and game-theoretical models~\cite{globerson06-icml,bruckner12,bulo17-tnnls}.
These approaches, which also encompass  \emph{adversarial training}~\cite{goodfellow15-iclr},
explicitly model the interactions between the classifier and the attacker to learn robust classifiers. The underlying idea is to incorporate knowledge of potential attacks during training.
The second family of defenses (complementary to the first) is based on the idea of rejecting samples that exhibit an outlying behavior with respect to unperturbed training data~\cite{melis17-vipar,bendale16-cvpr,crecchi19-esann,lu17-iccv,papernot18-arxiv}. 

In this work, we focus on defenses based on rejection mechanisms and try to improve their effectiveness. In fact, it has been shown that only relying upon the feature representation learned by the last network layer to reject adversarial examples is not sufficient~\cite{melis17-vipar,bendale16-cvpr}.  
In particular, it happens that adversarial examples become indistinguishable from samples of the target class at such a higher representation level even for small input perturbations.
To overcome this issue, we propose here a defense mechanism, named \emph{Deep Neural Rejection} (DNR), based on analyzing the representations of input samples at \emph{different} network layers, and on rejecting samples which exhibit anomalous behavior with respect to that observed from the training data at such layers (Sect.~\ref{sect:dnr}). 
With respect to similar approaches based on analyzing different network layers~\cite{lu17-iccv,crecchi19-esann}, our defense does not require generating adversarial examples during training, and it is thus less computationally demanding.

We evaluate our defense against an adaptive white-box attacker that is aware of the defense mechanism and tries to bypass it. 
To this end, we propose a novel gradient-based attack that accounts for the rejection mechanism and aims to craft adversarial examples that avoid it (Sect.~\ref{sect:attack}). 

It is worth remarking here that correctly evaluating a defense mechanism is a crucial point when proposing novel defenses against adversarial examples~\cite{athalye18,biggio18}.
The majority of previous work proposing defense methods against adversarial examples has only evaluated such defenses against previous attacks rather than against an ad-hoc attack crafted specifically against the proposed defense (see, \eg,~\cite{papernot16-distill,lu17-iccv,meng17-ccs} and all the other re-evaluated defenses in~\cite{athalye18,carlini17-aisec}).
The problem with these black-box and gray-box evaluations in which the attack is essentially unaware of the defense mechanism is that they are overly optimistic. 
It has indeed been shown afterwards that such defenses can be easily bypassed by simple modifications to the attack algorithm~\cite{athalye18,carlini17-sp,carlini17-aisec}. For instance, many defenses have been found to perform \emph{gradient obfuscation}, \ie, they learn functions which are harder to optimize for gradient-based attacks; however, they can be easily bypassed by constructing a smoother, differentiable approximation of their function, \eg, via learning a surrogate model~\cite{biggio13-ecml,biggio18,russu16-aisec,papernot17-asiaccs,demontis19-usenix,melis18-eusipco} or replacing network layers which obfuscate gradients with smoother mappings~\cite{athalye18,carlini17-sp,carlini17-aisec}.
In our case, an attack that is unaware of the defense mechanism may tend to craft adversarial examples in areas of the input space which are assigned to the rejection class; thus, such attacks, as well as previously-proposed ones, may rarely bypass our defense.
For this reason, we believe that our adaptive white-box attack, along with the security evaluation methodology adopted in this work, provide another significant contribution to the state of the art related to the problem of properly evaluating defenses against adversarial examples.

The security evaluation methodology advocated in~\cite{biggio14-tkde,biggio18,athalye18}, which we also adopt in this work, consists of evaluating the accuracy of the system against attacks crafted with an increasing amount of perturbation. The corresponding \emph{security evaluation curve}~\cite{biggio18} shows how gracefully the performance decreases while the attack increases in strength, up to the point where the defense reaches zero accuracy. This is another important phenomenon to be observed, since any defense against test-time evasion attacks \emph{has to fail} when the perturbation is sufficiently large (or, even better, unbounded); in fact, in the unbounded case, the attacker can ideally replace the source sample with any other sample from another class~\cite{athalye18}.
If accuracy under attack does not reach zero for very large perturbations, then it may be that the attack algorithm fails to find a good optimum (\ie, a good adversarial example). This in turn means that we are probably providing an optimistic evaluation of the defense.
As suggested in~\cite{athalye18}, the purpose of a security evaluation should not be to show which attacks the defense withstands to, but rather to show when the defense fails. If one shows that larger perturbations that may compromise the content of the input samples and its nature (\ie, its true label) are required to break the defense, then we can retain the defense mechanism to be sufficiently robust. Another relevant point is to show that such a \emph{breakdown} point occurs at a larger perturbation than that exhibited by competing defenses, to show that the proposed defense is more robust than previously-proposed ones.

The empirical evaluation reported in Sect.~\ref{sect:exp}, using both MNIST handwritten digits and CIFAR10 images, provides consistent results with the aforementioned aspects. First, it shows that our adaptive white-box attack is able to break our defensive method at larger perturbations. Second, it shows that our method improves the performance of competing rejection mechanisms which only leverage the deep representation learned at the output network layer.
We thus believe that our analysis unveils a promising way of defending against adversarial examples.

We conclude the paper by discussing related work (Sect.~\ref{sect:related}), the main contributions of this work and its limitations, along with promising future research directions (Sect.~\ref{sect:conc}).

\section{Deep Neural Rejection}
\label{sect:dnr}

The underlying idea of our DNR method is to estimate the distribution of unperturbed training points at different network layers, and reject anomalous samples that may be incurred at test time, including adversarial examples. The architecture of DNR is shown in Fig.~\ref{fig:method}. 

Before delving into the details of our method, let us introduce some notation. We denote the prediction function of a deep neural network with $f : \set X \mapsto \set Y$, where $\set X \subseteq \mathbb R^d$ is the $d$-dimensional space of input samples (\eg, image pixels) and $\set Y \subseteq \mathbb R^c$ is the space of the output predictions (\ie, the estimated confidence values for each class), being $c$ the number of classes.
If we assume that the network consists of $m$ layers, then the prediction function $f$ can be rewritten to make this explicit as: $f(\phi_1( \phi_2( \ldots \phi_m( \vct x; \vct w_m) ; \vct w_2); \vct w_1)$, where $\phi_1$ and $\phi_m$ denote the mapping function learned respectively by the output and the input layer, and $\vct w_1$ and $\vct w_m$ are their weight parameters (learned during training).

For our defense mechanism to work, one has first to select a set of network layers; \eg, in Fig.~\ref{fig:method} we select the outer layers $\phi_1$, $\phi_2$ and $\phi_3$. 
Let us assume that the representation of the input sample $\vct x$ at level $\phi_i$ is $\vct z_i$.
Then, on each of these selected representations, DNR learns an SVM with the RBF kernel $g_i(\vct z_i)$, trying to correctly predict the input sample. The confidence values on the $c$ classes provided by this classifier are then concatenated with those provided by the other base SVMs, and used to train a combiner, using again an RBF SVM.\footnote{Validation samples should be used to train the combiner here and avoid overfitting, as suggested by stacked generalization~\cite{wolpert92}.} The combiner will output predictions $s_1, \ldots, s_c$ for the $c$ classes, but will reject samples if the maximum confidence score $\max_{k=1, \ldots, c} s_k$ is not higher than a rejection threshold $\theta$.
This decision rule can be compactly represented as:
$\argmax_{k=0,\ldots,c} s_k(\vct x)$, where we define an additional, \emph{constant} output $s_0(\vct x)=\theta$ for the rejection class. According to this rule, if $s_0(\vct x) = \theta$ is the highest value in the set, the sample is rejected; otherwise, it is assigned to the class exhibiting the larger confidence value.

As proposed in~\cite{melis17-vipar}, we use an RBF SVM here to ensure that the confidence values $s_1, \ldots, s_c$, for each given class, decrease while $\vct x$ moves further away from regions of the feature space which are densely populated by training samples of that class. 
This property, named \emph{compact abating probability} in open-set problems~\cite{scheirer14,bendale16-cvpr}, is a desirable property to easily implement a \emph{distance-based rejection} mechanism as the one required in our case to detect outlying samples.
With respect to~\cite{melis17-vipar}, we train this combiner on top of other base classifiers rather than only on the representation learned by the last network layer, to further improve the detection of adversarial examples. For this reason, in the following we refer to the approach by Melis~\etal~\cite{melis17-vipar}, rejecting samples based only on their representation at the last layer, as \emph{Neural Rejection} (NR); and to ours, exploiting also representations from other layers, as \emph{Deep Neural Rejection} (DNR).

\section{Attacking Deep Neural Rejection}
\label{sect:attack}

\begin{figure}[t]    
	\centering
	\includegraphics[height=.21\textwidth]{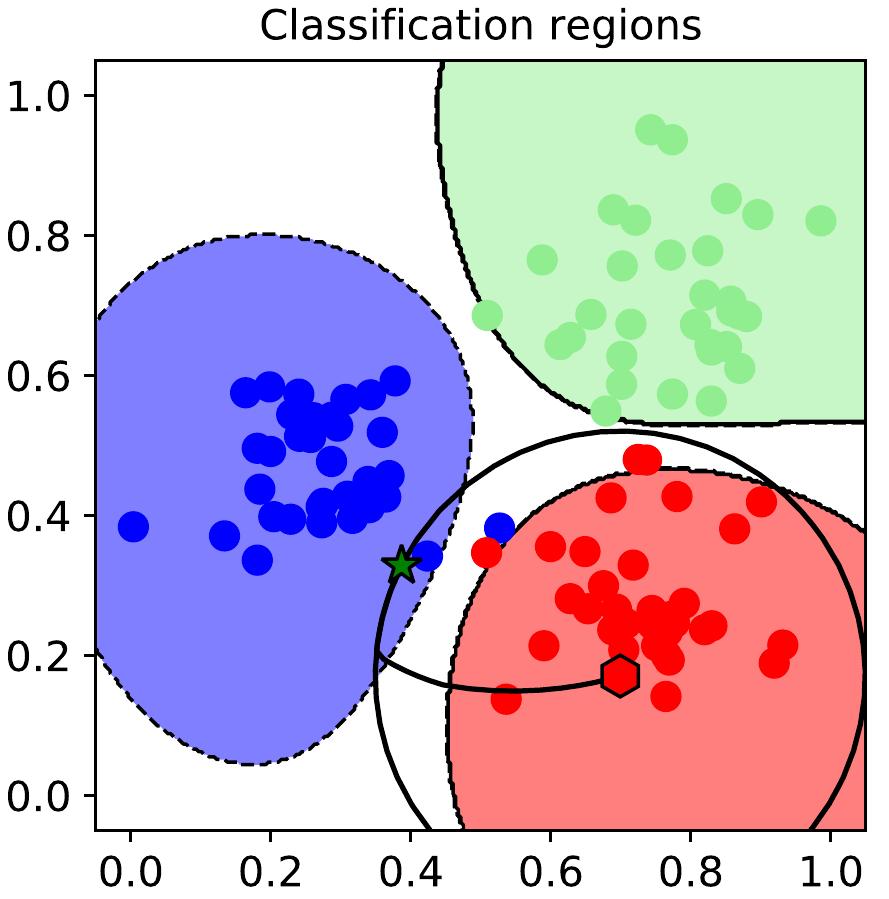}
	\includegraphics[height=.21\textwidth]{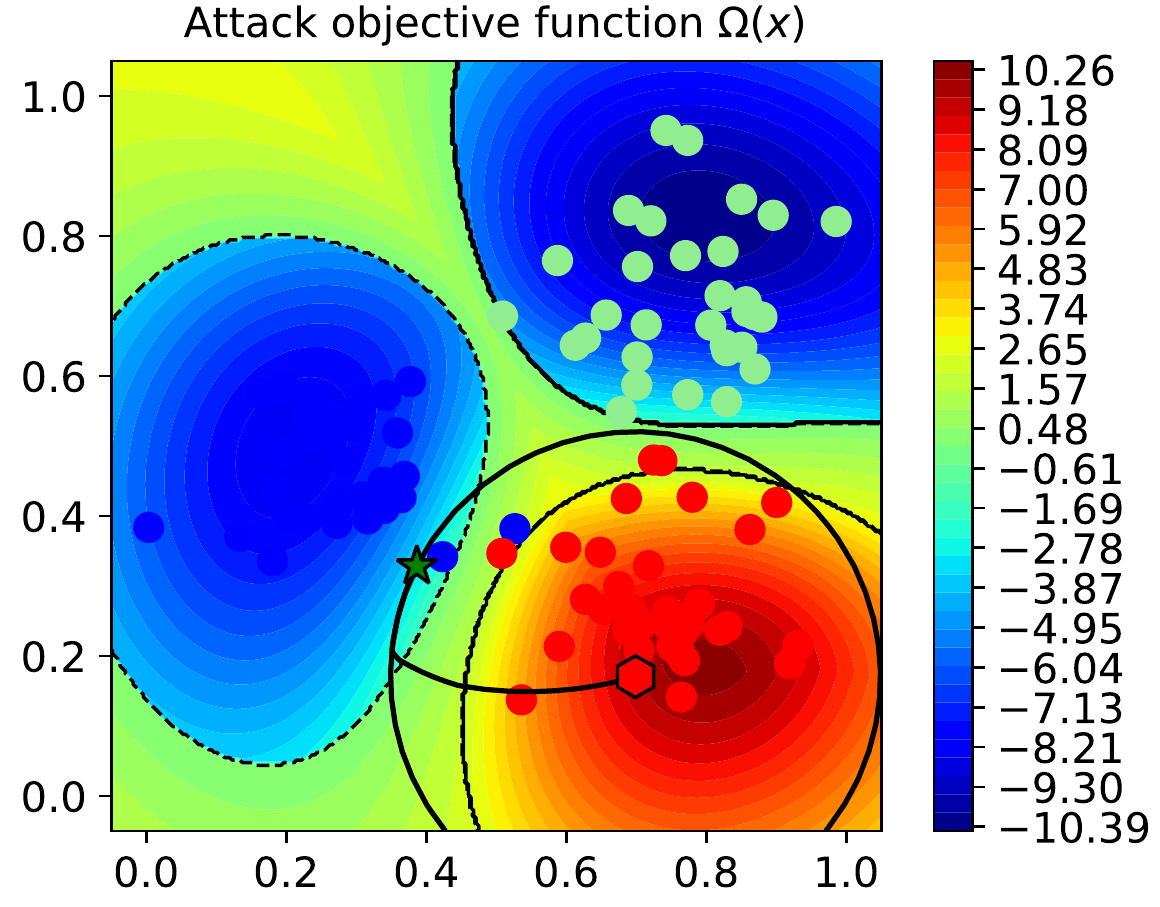}
	\caption{Our defense-aware attack against an RBF SVM with rejection, on a 3-class bi-dimensional classification problem. The initial sample $x_0$ and the adversarial example $x^\star$ are respectively represented as a red exagon and a green star, while the $\ell_2$-norm perturbation constraint $\|x_0-x^\prime\|_2 \leq \varepsilon$ is shown as a black circle.
		The left plot shows the decision region of each class, along with the reject region (in white). The right plot shows the values of the attack objective $\Omega(\vct x)$ (in colors), which correctly enforces our attacks to avoid the reject region.}
	\label{fig:indiscriminate_evasion}
\end{figure}

To properly evaluate security, or \emph{adversarial robustness}, of rejection-based defenses against adaptive white-box adversarial examples, we propose the following. Given a source sample $\vct x$ and a maximum-allowed $\varepsilon$-sized perturbation, the attacker can optimize a defense-aware adversarial example $\vct x^\star$ by solving the following constrained optimization problem:
\begin{eqnarray}
\label{eq:obj}	
\vct x^\star = \argmin_{\vct x^\prime : \| \vct x- \vct x^\prime\| \leq \varepsilon} \Omega(\vct x) 
\end{eqnarray}
where:
\begin{eqnarray}
\label{eq:onega}	
\Omega(\vct x) = s_y( \vct x^\prime) - \max_{j \not \in \{0, y\} } s_j(\vct{x^\prime}) \, ,  
\end{eqnarray}

where $\| \vct x- \vct x^\prime\| \leq \varepsilon$ is an $\ell_p$-norm constraint (typical norms used for crafting adversarial examples are $\ell_1$, $\ell_2$ and $\ell_\infty$, for which efficient projection algorithms exist~\cite{duchi08}), $y \in \set Y = \{1, \ldots, c\}$ is the true class, and $0$ is the rejection class. In practice, the attacker minimizes the output of the true class, while maximizing the output of the competing class (excluding the reject class) to achieve (untargeted) evasion. 
This amounts to performing a strong maximum-confidence evasion attack (rather than searching for a minimum-distance adversarial example). We refer the reader to~\cite{biggio13-ecml,biggio18,demontis19-usenix} for a more detailed discussion on such topic.
While we focus here on untargeted (error-generic) attacks, our formulation can be extended to account for targeted (error-specific) evasion as also done in~\cite{melis17-vipar}.

The optimization problem in Eq.~\eqref{eq:obj} can be solved through a standard projected gradient descent (PGD) algorithm, as given in Algorithm~\ref{alg:evasion}. In our experiments, we consider a variable step size $\eta$ (by doubling the initial step size for ten times), and select the point $\vct x^\prime$ minimizing the objective at each update step. This allows our attack to escape local minima which may hinder the optimization process and, consequently, it allows us to obtain a more reliable security evaluation of the proposed detection method.

\begin{algorithm}[h]
	\caption{PGD-based Maximum-confidence Adversarial Examples}
	\label{alg:evasion}
	\begin{algorithmic}[1]
		\Require $\vct x_{0}$: the input sample;  $\eta$: the step size; $\Pi$: a projection operator on the $\ell_p$-norm constraint $\| \vct x_0- \vct x^\prime\| \leq \varepsilon$; $t > 0$: a small positive number to ensure convergence. 
		\Ensure $\vct x^{\prime}$: the adversarial example.
		\State $\vct x^{\prime} \gets \vct x_{0}$
		\Repeat
		\State $\vct x \gets \vct x^{\prime}$
		\State $ \vct x^\prime \gets  \Pi \left ( \vct x -  \eta \nabla \Omega(\vct x) \right ) $
		\Until{$ | \Omega (\vct x^\prime) - \Omega (\vct x) | \le t $}
		\State \Return $\vct x^\prime$
	\end{algorithmic}
\end{algorithm}

In Fig.~\ref{fig:indiscriminate_evasion} we report an example on a bi-dimensional toy problem to show how our defense-aware attack works against a rejection-based defense mechanism.

\section{Experimental Analysis}
\label{sect:exp}

In this section, we evaluate the security of the proposed DNR method against adaptive, defense-aware adversarial examples. We consider two common computer-vision benchmarks for this task, \ie, handwritten digit recognition (MNIST data) and image classification (CIFAR10 data). Our goal is to investigate whether and to which extent DNR can improve security against adversarial examples; in particular, compared to the previously-proposed neural rejection (NR) defense (which only leverages the feature representation learned at the last network layer to reject adversarial examples)~\cite{melis17-vipar}.
All the experiments presented in this section are based on the open-source Python library \texttt{secml}~\cite{secml}, which we plan to extend in the near future to include an implementation of both DNR and NR.

\subsection{Experimental Setup}
We discuss here the experimental setup used to evaluate our defense mechanism.

\begin{table}[!htb]
	\caption{Model architecture of the MNIST network}
	\label{tab:architecture_mnist}
	\centering
	\begin{tabular}{lc}
		\toprule
		Layer Type & Dimension\\
		\midrule
		Conv. + ReLU & 32 filters (3x3) \\
		Conv. + ReLU & 32 filters (3x3) \\
		Max Pooling & 2×2 \\
		Conv. + ReLU & 64 filters (3x3) \\
		Conv. + ReLU & 64 filters (3x3) \\
		Max Pooling &  2×2 \\
		Fully Connected + ReLU & 200 units \\
		Fully Connected + ReLU & 200 units \\
		Softmax  & 10 units \\
		\bottomrule
	\end{tabular}
\end{table}

\begin{table}[!htb]
	\label{tab:architecture_cifar}
	\centering
	\caption{Model architecture of the CIFAR10 network}
	\begin{tabular}{lc}
		\toprule
		Layer Type & Dimension \\
		\midrule
		Conv. + Batch Norm. + ReLU & 64 filters (3x3) \\
		Conv. + Batch Norm. + ReLU & 64 filters (3x3) \\
		Max Pooling + Dropout ($p=0.1$) & 2x2 \\
		Conv. + Batch Norm. + ReLU & 128 filters (3x3) \\
		Conv. + Batch Norm. + ReLU & 128 filters (3x3) \\
		Max Pooling + Dropout ($p=0.2$) & 2x2 \\
		Conv. + Batch Norm. + ReLU & 256 filters (3x3) \\
		Conv. + Batch Norm. + ReLU & 256 filters (3x3) \\
		Max Pooling + Dropout ($p=0.3$) & 2x2 \\
		Conv. + Batch Norm. + ReLU & 512 filters (3x3) \\
		Max Pooling + Dropout ($p=0.4$) & 2x2 \\
		Fully Connected & 512 units \\
		Softmax & 10 units \\
		\bottomrule
	\end{tabular}
\end{table}

\begin{table}[h]
	\centering
	\caption{Parameters used to train MNIST and CIFAR10 networks}
	\label{tab:params}
	\begin{tabular}{lll}
		\toprule
		Parameter & MNIST Model & CIFAR Model \\
		\midrule
		Learning Rate & 0.1 & 0.01 \\
		Momentum & 0.9 & 0.9 \\
		Dropout & 0.5 & (see table~\ref{tab:architecture_cifar}) \\
		Batch Size & 128 & 100 \\
		Epochs & 50 & 75 \\
		\bottomrule
	\end{tabular}
\end{table}

\myparagraph{Datasets.} As mentioned before, we run experiments on MNIST and CIFAR10 data. MNIST handwritten digit data consists of 60,000 training and 10,000 test gray-scale 28x28 images. CIFAR10 consists of 50,000 training and 10,000 test RGB 32x32 images. We normalized the images of both datasets in [0,1] by simply dividing the input pixel values by 255. 

\myparagraph{Train-test splits.} We average the results on five different runs. In each run, we consider 10,000 training samples and 1,000 test samples, randomly drawn from the corresponding datasets. To avoid overfitting, we train the DNR combiner on the outputs of the base SVMs computed on a separate validation set, using a procedure known as \emph{stacked generalization}~\cite{wolpert92}.
We use a 3-fold cross-validation procedure to subdivide the training dataset into three folds. For three times, we learn the base SVMs on two folds and classify the remaining (validation) fold. We then concatenate the predicted values for each validation fold and use such values to train the combiner. 
The deep neural networks (DNNs) used in our experiments are pre-trained on a training dataset (different from the ones that we use to train the SVMs) of 30,000 and 40,000 training samples, respectively, for MNIST and CIFAR10.

\myparagraph{Classifiers.} We compare the DNR approach (which implements rejection here based on the representations learned by three different network layers) against an undefended DNN (without any rejection mechanism) and against the NR defense by Melis~\etal~\cite{melis17-vipar} (which implements rejection on top of the representation learned by the output network layer). To implement the undefended DNNs for the MNIST dataset, we used the same architecture suggested by Carlini~\etal~\cite{carlini17-sp}. For CIFAR10, instead, we considered a lightweight network that, despite its size, allows obtaining high performances. 
The two considered architectures are shown in Table~\ref{tab:architecture_mnist} and Table~\ref{tab:architecture_cifar}, whereas table~\ref{tab:params} shows the model parameters that we used to train the overmentioned architectures. The three layers considered by the DNR classifier are the last three layers for the network trained on MNIST, and the last layer plus the last batch norm layer and the second to last max-pooling layer for the one trained on CIFAR10 (chosen to obtain a reasonable amount of features).

\begin{figure*}[t]    
	\centering
	\includegraphics[width=.42\textwidth]{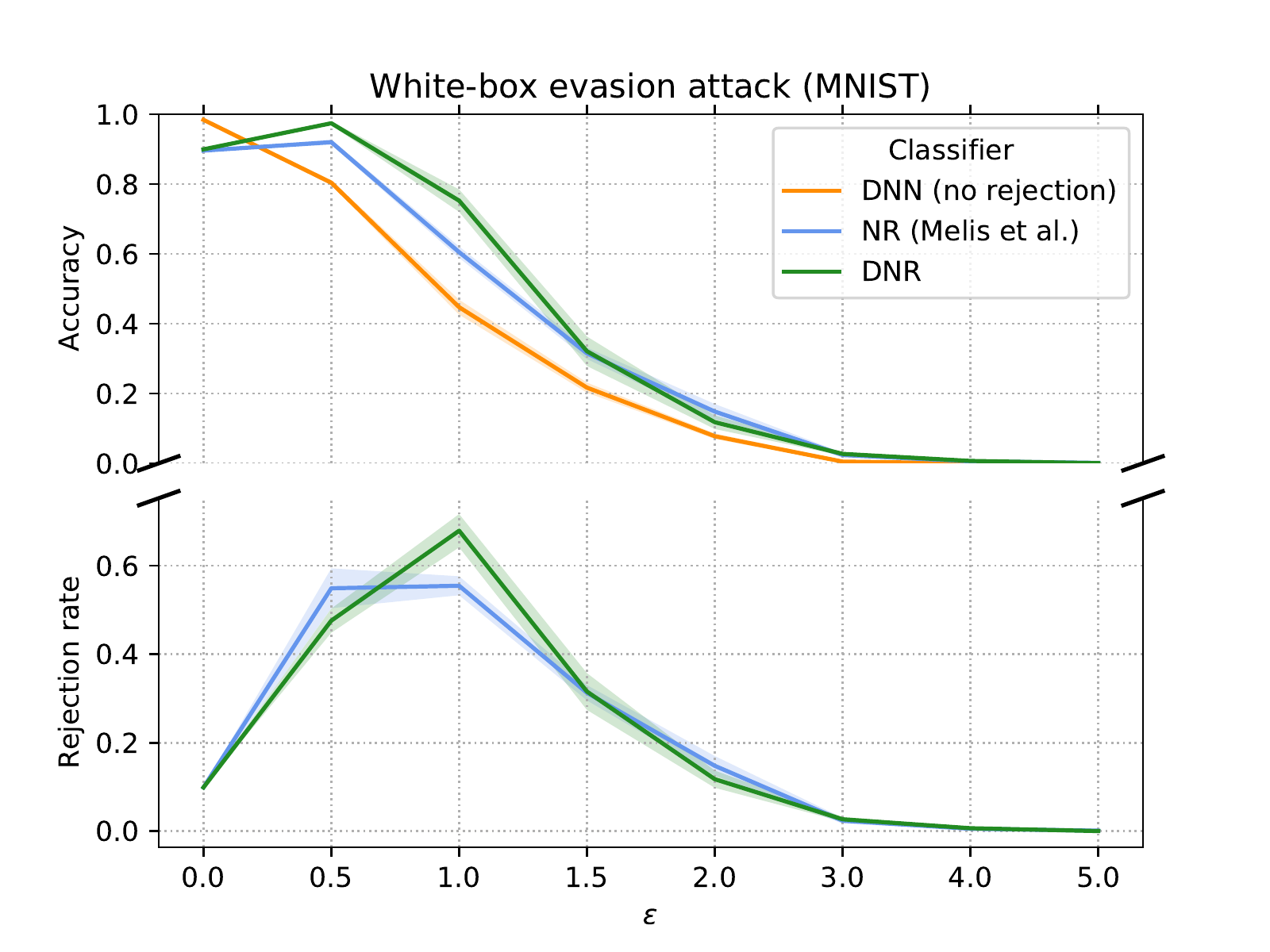}
	\includegraphics[width=.42\textwidth]{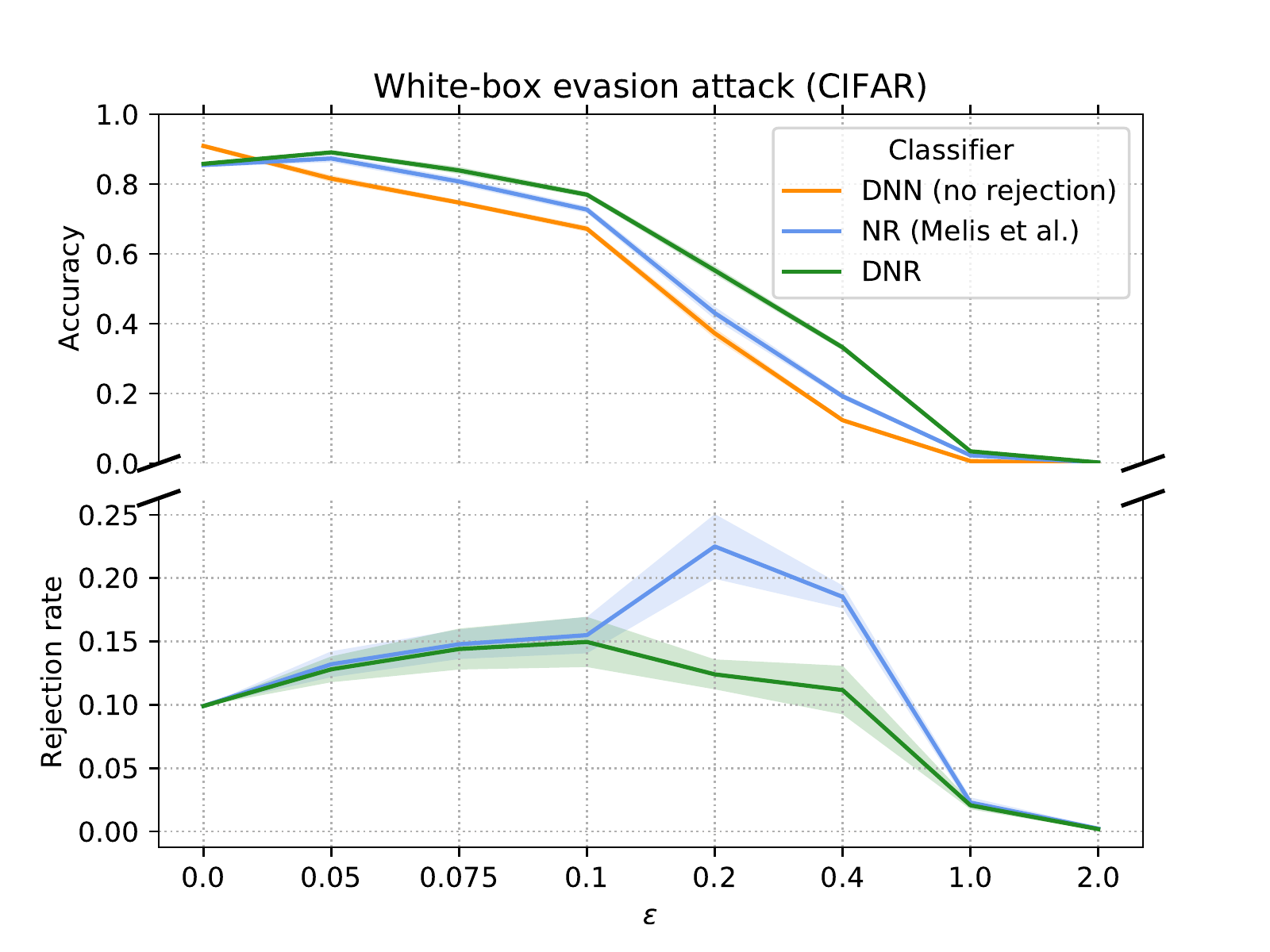}
	\caption{Security evaluation curves for MNIST (left) and CIFAR10 (right) data, reporting mean accuracy (and standard deviation) against $\varepsilon$-sized attacks.}
	\label{fig:sec-eval}
\end{figure*}

\begin{figure*}[t]    
	\centering
	\includegraphics[width=.42\textwidth]{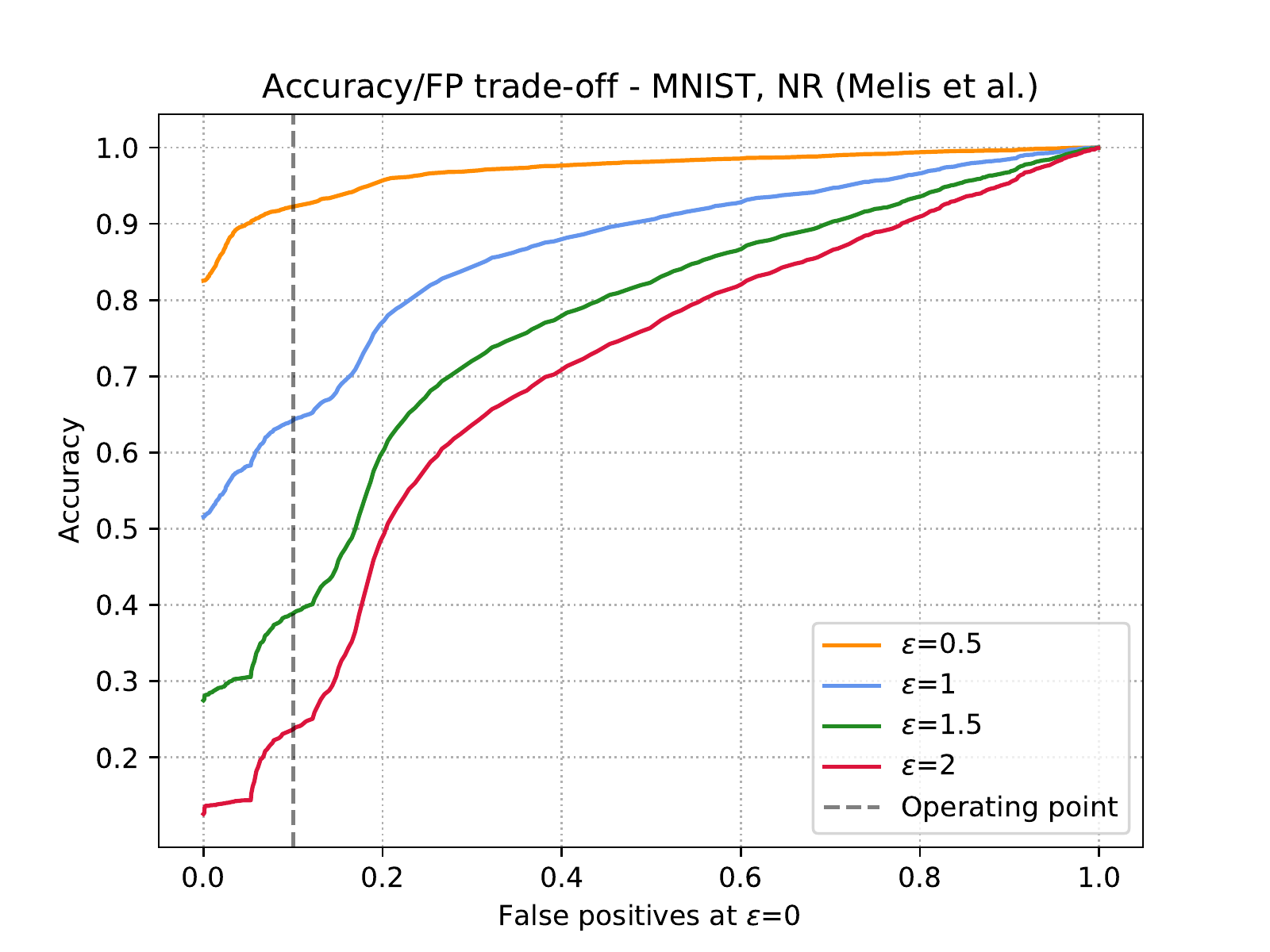}
	\includegraphics[width=.42\textwidth]{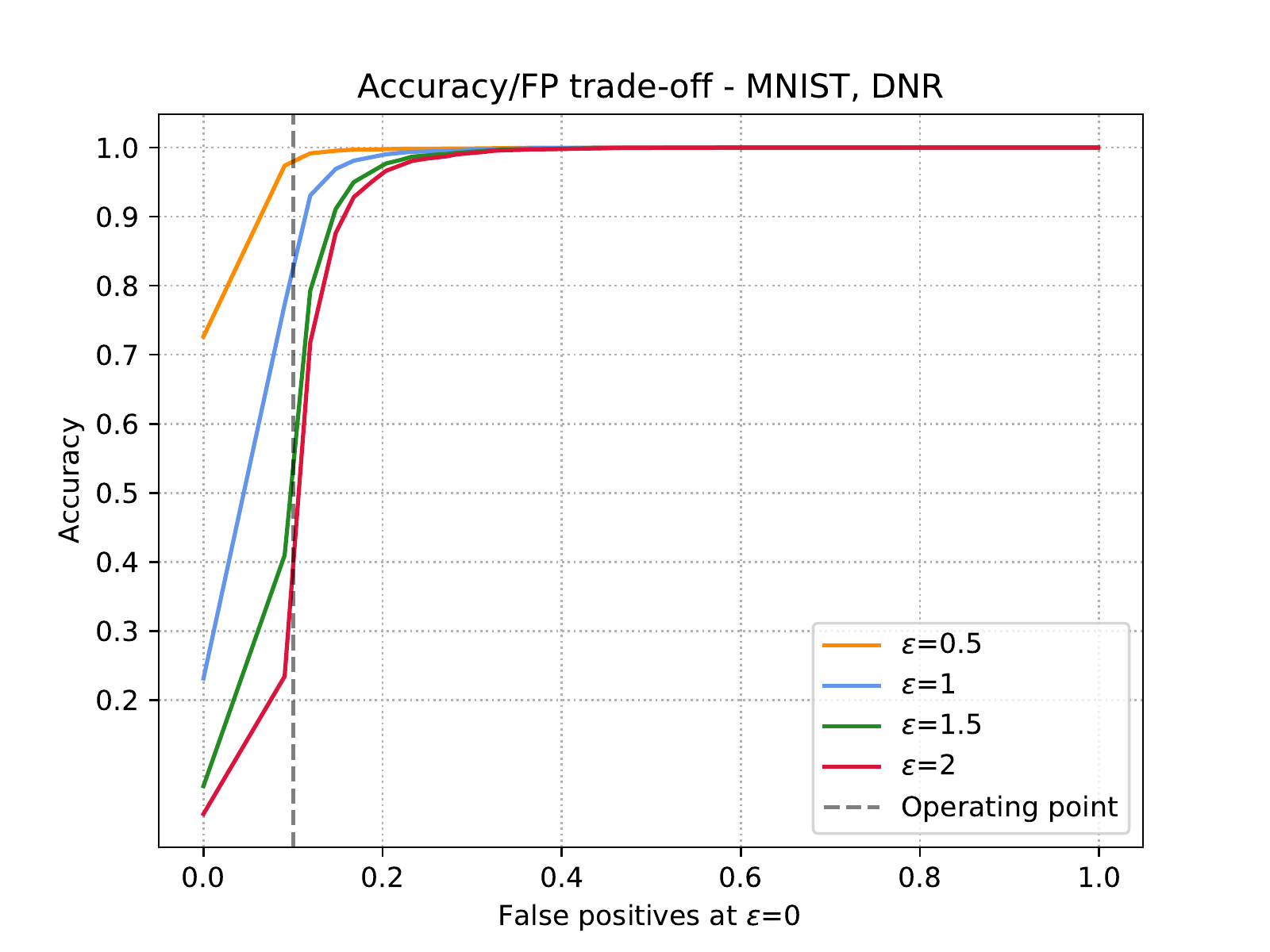}
	\includegraphics[width=.42\textwidth]{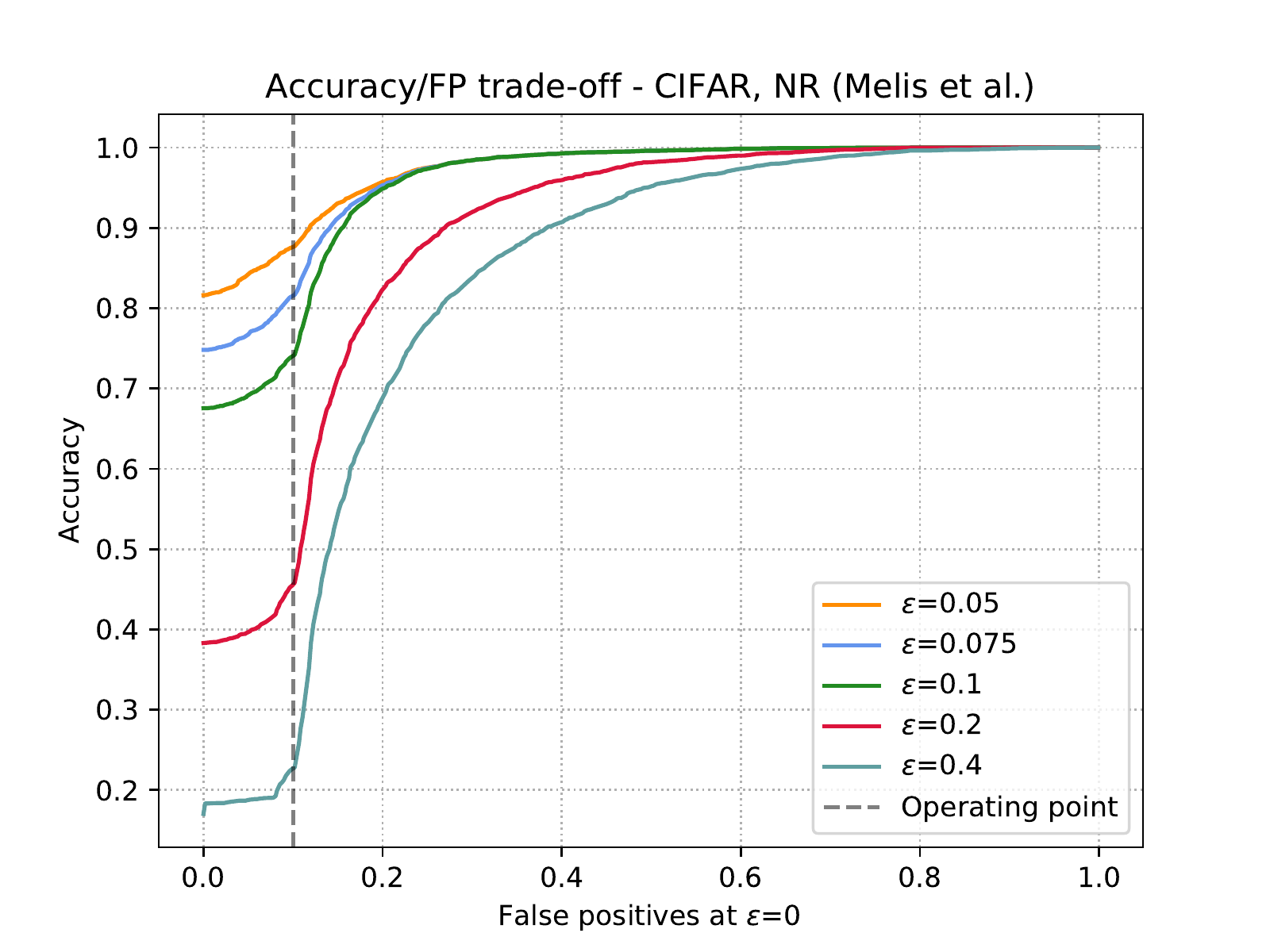}
	\includegraphics[width=.42\textwidth]{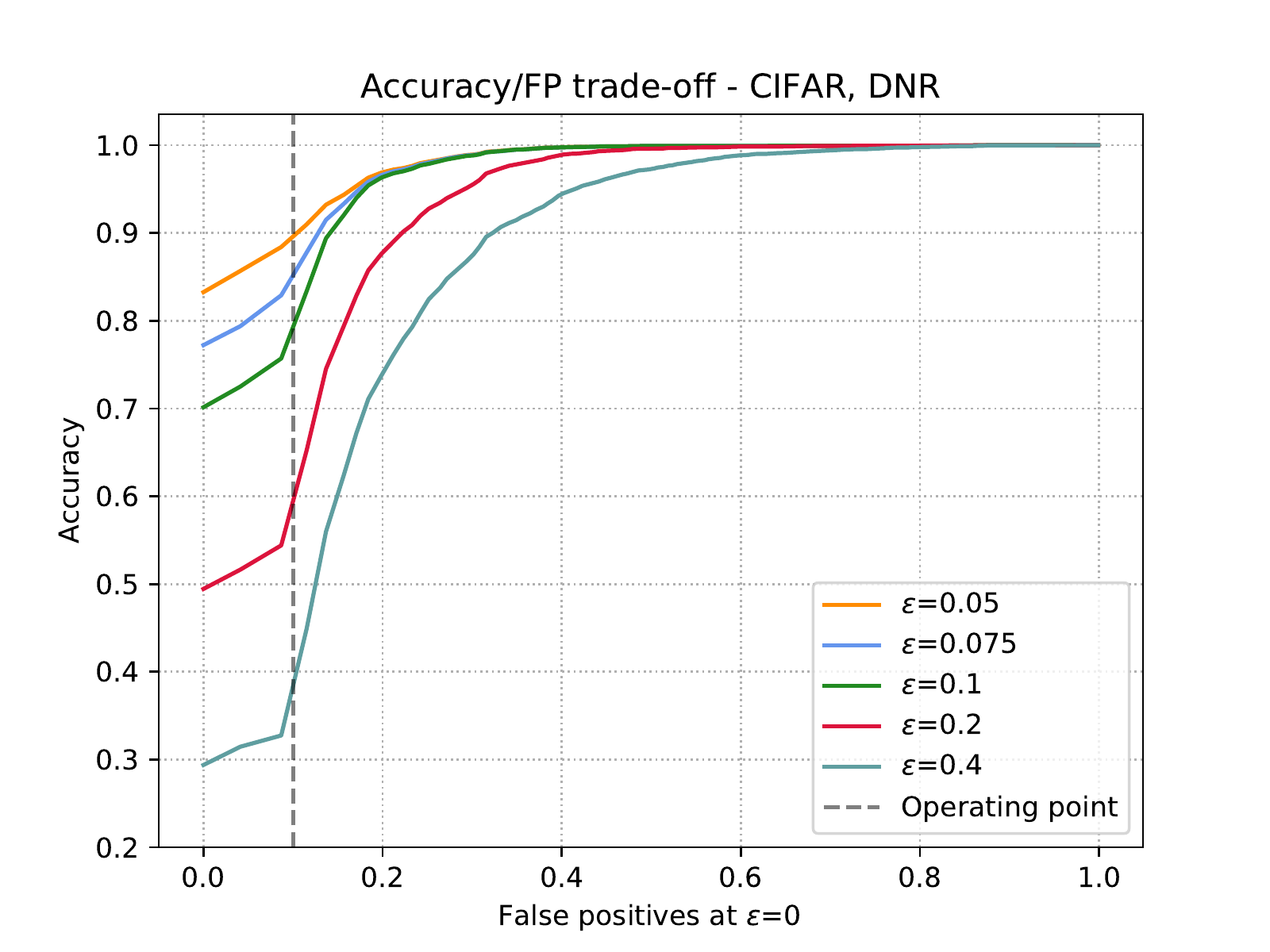}
	\caption{
		Influence of the rejection threshold $\theta$ on classifier accuracy under attack (y-axis) vs false rejection rate (\ie, fraction of wrongly-rejected unperturbed samples) on MNIST (top) and CIFAR10 (bottom) for NR (left) and DNR (right), for different $\varepsilon$-sized attacks. The dashed line highlights the performance at 10\% false rejection rate (\ie, the operating point used in our experiments).}
	\label{fig:thresh-selection}
\end{figure*}

\myparagraph{Security Evaluation.} We compare these classifiers in terms of their security evaluation curves~\cite{biggio18}, reporting classification accuracy against an increasing $\ell_2$-norm perturbation size $\varepsilon$, used to perturb all test samples. In particular, classification accuracy is computed as follows:
\begin{itemize}
	\item in the absence of adversarial perturbation (\ie, for $\varepsilon=0$), classification accuracy is computed as usual, but considering rejects as errors;
	\item in the presence of adversarial perturbation (\ie, for $\varepsilon>0$), all test samples become adversarial examples, and we consider them correctly classified if they are assigned either to the rejection class or to their original class (which typically happens when the perturbation is too small to cause a misclassification).
\end{itemize}
For DNR and NR, we also report the rejection rates, computed by dividing the number of rejected samples by the number of test samples.
Note that the difference between accuracy and rejection rate at each $\varepsilon>0$ corresponds to the fraction of adversarial examples which are not rejected but still correctly assigned to their original class.
Accordingly, under this setting, classifiers exhibiting higher accuracies under attack ($\varepsilon>0$) can be retained more robust.

\myparagraph{Parameter setting.} We use a 5-fold cross-validation procedure to select the hyperparameters that maximize classification accuracy on the unperturbed training data, and set the rejection threshold $\theta$ for NR and DNR to reject 10\% of the samples when no attack is performed (at $\varepsilon=0$).

\begin{figure}[t]	
	\centering
	\includegraphics[width=.40\textwidth]{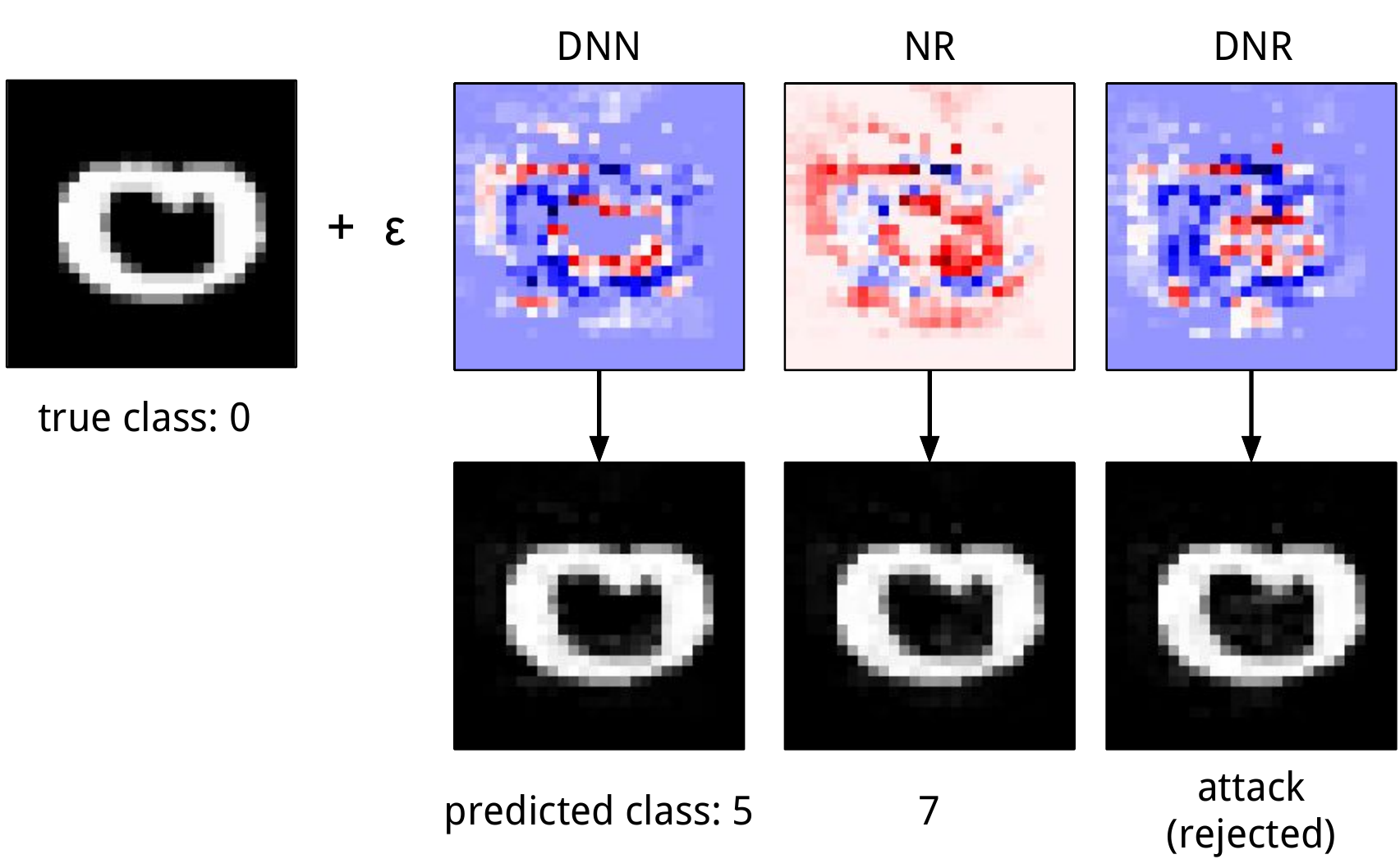} 
	\vspace{0.5cm}
	\includegraphics[width=.40\textwidth]{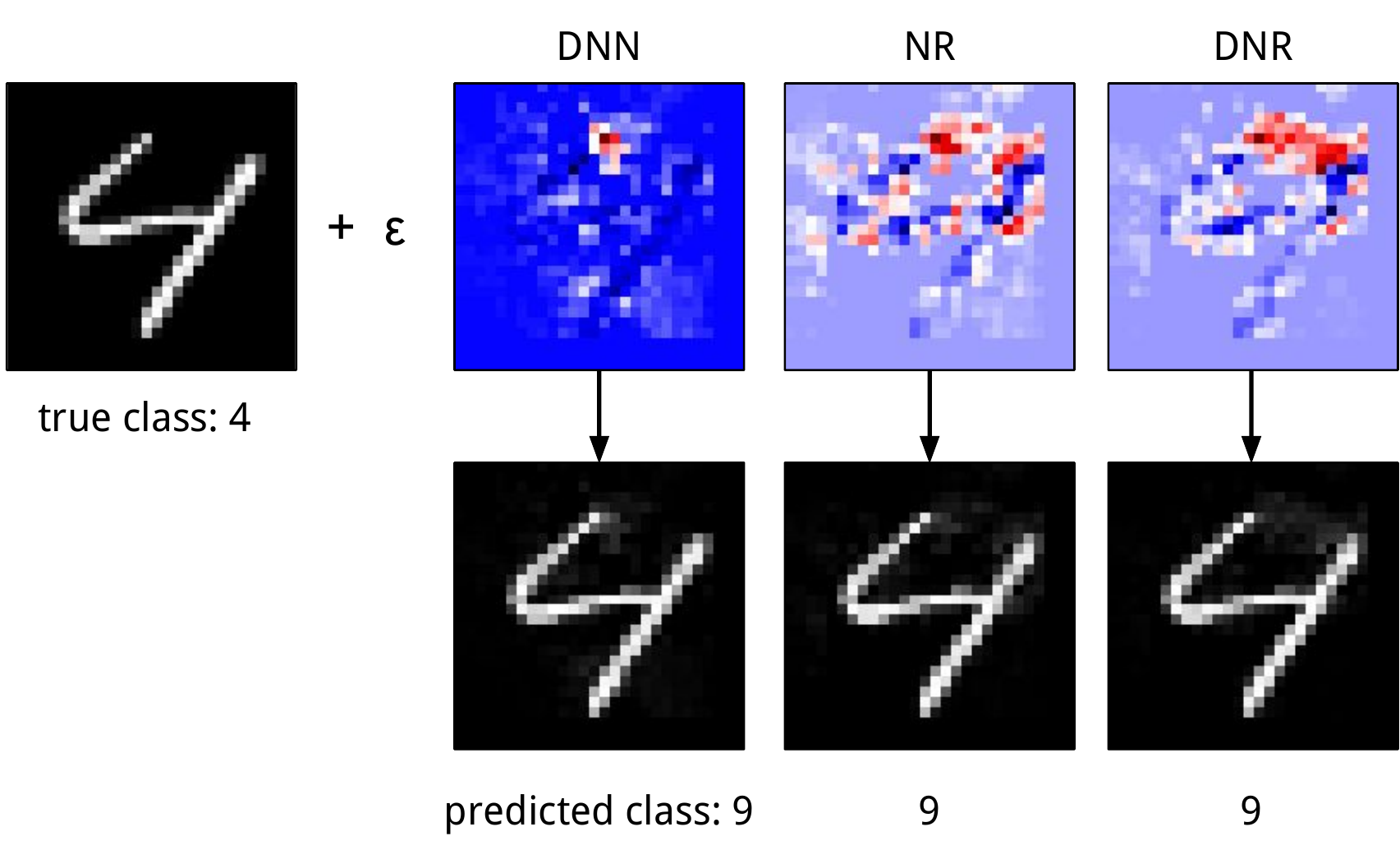} 
	\caption{Adversarial examples computed on the MNIST data to evade DNN, NR and DNR. The source image is reported on the left, followed by the (magnified) adversarial perturbation crafted  with $\varepsilon = 1$ against each classifier, and the resulting adversarial examples. We remind the reader that the attacks considered in this work are untargeted, i.e., they succeed when the attack sample is not correctly assigned to its true class.}
	\label{fig:mnist_adversarial_digits}
\end{figure}

\subsection{Experimental Results}

The results are reported in Fig.~\ref{fig:sec-eval}.
In the absence of attack ($\varepsilon=0$), the undefended DNNs slightly outperform NR and DNR, since the latter wrongly reject also some unperturbed samples.

\begin{figure}[t]	
	\centering
	\includegraphics[width=.40\textwidth]{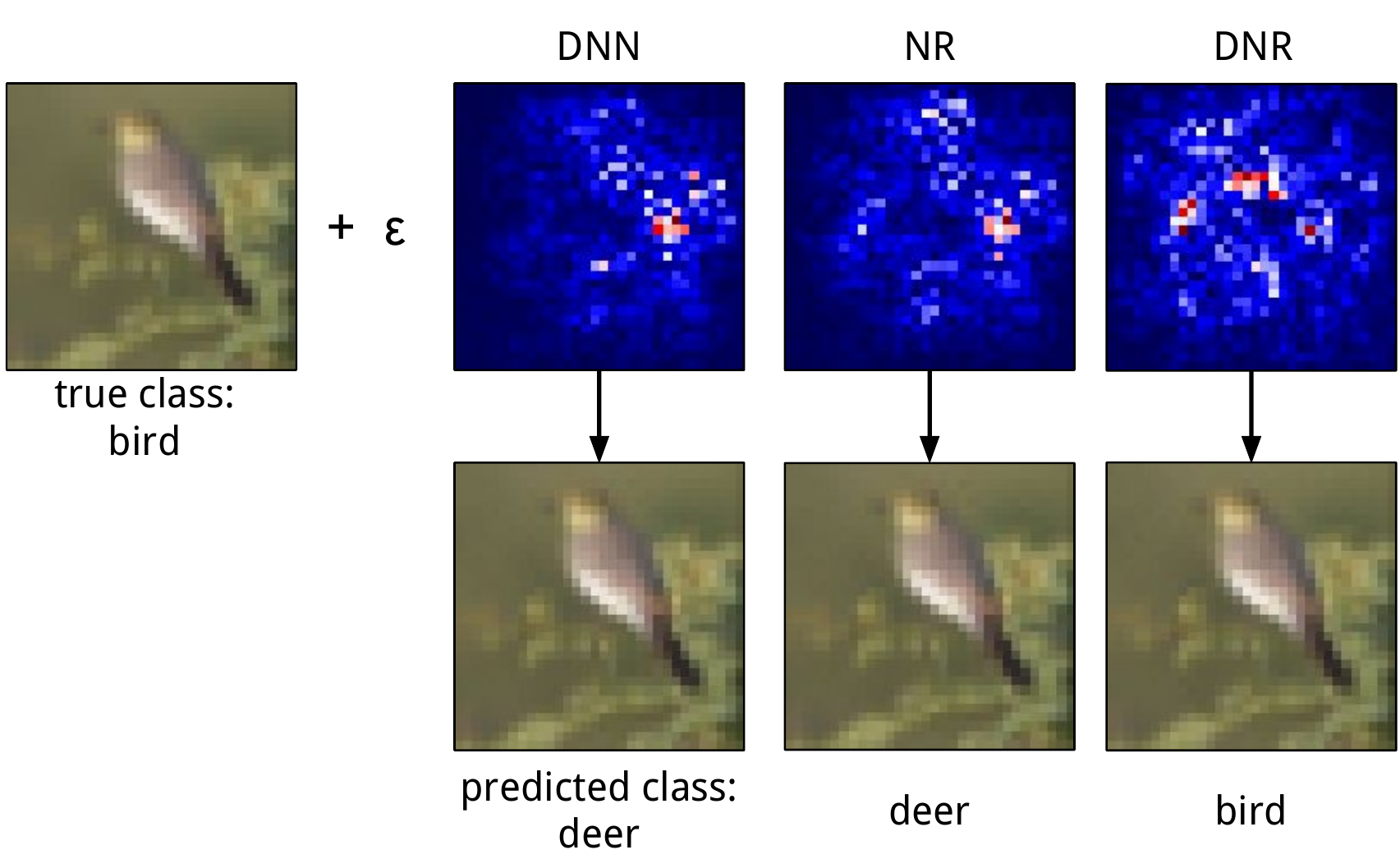} 
	\vspace{0.5cm}
	\includegraphics[width=.40\textwidth]{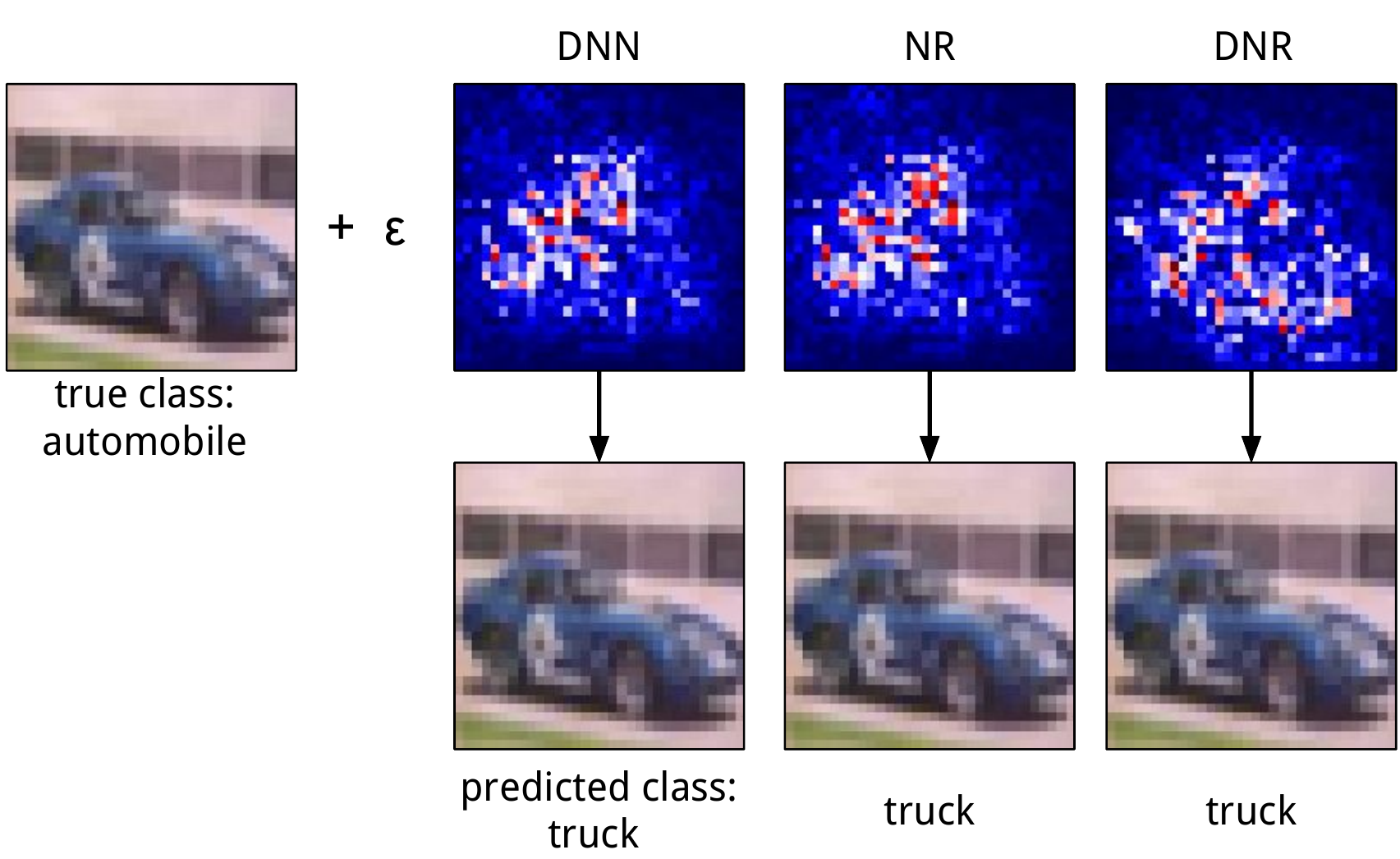} 
	\caption{Adversarial examples computed on the CIFAR10 dataset adding a perturbation computed with $\varepsilon = 0.2$. See the caption of Fig.\ref{fig:mnist_adversarial_digits} for further details.}
	\label{fig:cifar_adversarial_digits}
\end{figure}

Under attack, ($\varepsilon>0$), when the amount of injected perturbation is exiguous, the rejection rate of both NR and DNR increases jointly with $\varepsilon$, as the adversarial examples are located far from the rest of the training classes in the representation space (\ie, the intermediate representations learned by the neural network). For larger $\varepsilon$, both NR and DNR can no longer correctly detect the adversarial examples, as they tend to become indistinguishable from the rest of the training samples (in the representation space in which NR and DNR operate).
Both defenses outperform the undefended DNNs on the adversarial samples, and DNR slightly outperforms NR, exhibiting a more graceful decrease in performance. 
Although NR tends to reject more samples for $\varepsilon \in [0.1, 1]$ on CIFAR and for $\varepsilon=0.5$ on MNIST, its accuracy is lower than DNR. The reason is that DNR remains more accurate than NR when classifying samples that are not rejected. 
This also means that DNR provides tighter boundaries closer to the training classes than NR, thanks to the exploitation of lower-level network representations, which makes the corresponding defended classifier more difficult to evade. In Fig.~\ref{fig:mnist_adversarial_digits} and Fig.~\ref{fig:cifar_adversarial_digits}, we show some adversarial examples computed respectively on the MNIST and CIFAR10 datasets against the considered classifiers.

Finally, in Fig.~\ref{fig:thresh-selection} we show how the selection of the rejection threshold $\theta$ allows us to trade security against adversarial examples (\ie, accuracy on the y axis) for a more accurate classifier on the unperturbed samples (reported in terms of the rejection rate of unperturbed samples on the x axis).
In particular, increasing (decreasing) the rejection threshold amounts to increasing (decreasing) the fraction of correcly-detected adversarial examples, and to increasing (decreasing) the rejection rate when no attack is performed.

\section{Related Work} 
\label{sect:related}

Different approaches have been recently proposed to perform rejection of samples that are outside of the training data distribution~\cite{thulasidasan19-iclr-rejected,geifman19-iclm,guyon17-nips}. For example, Thulasidasan~\etal~\cite{thulasidasan19-iclr-rejected} and Geifman~\etal~\cite{geifman19-iclm} have proposed novel loss functions accounting for rejection of inputs on which the classifier is not sufficiently confident. Guyon~\etal~\cite{guyon17-nips} have proposed a method that allows the system designer to set the desired risk level by adding a rejection mechanism to a pre-trained neural network architecture. These approaches have, however, not been originally tested against adversarial examples, and it is thus of interest to assess their performance under attack in future work, also in comparison to our proposal.

Even if the majority of approaches implementing rejection or abstaining classifiers have not considered the problem of defending against adversarial examples, some recent work has explored this direction too~\cite{bendale16-cvpr,melis17-vipar}. 
Nevertheless, with respect to the approach proposed in this work, they have only considered the output of the last network layer and perform rejection based solely on that specific feature representation. 
In particular, Bendale~and~Boult~\cite{bendale16-cvpr} have proposed a rejection mechanism based on reducing the open-set risk in the feature space of the activation vectors extracted from the last layer of the network, while Melis~\etal~\cite{melis17-vipar} have applied a threshold on the output of an RBF SVM classifier. Despite these differences, the rationale of the two approaches is quite similar and resembles the older idea of distance-based rejection.

Few approaches have considered a multi-layer detection scheme similar to that envisioned in our work~\cite{lu17-iccv,Carrara18-ECCV-ws,crecchi19-esann,pang18-iclm,papernot18-arxiv}. However, most of these approaches require generating adversarial examples at training time, which is computationally intensive, especially for high-dimensional problems and large datasets~\cite{lu17-iccv,Carrara18-ECCV-ws,crecchi19-esann,pang18-iclm}. Finding a methodology to tune the hyperparameters for generating the attack samples is also an open research challenge.
Finally, even though the DkNN approach by Papernot~\etal~\cite{papernot18-arxiv} does not require generating adversarial examples at training time, it requires computing the distance of each test sample against all the training points at different network layer representations, which again raises scalability issues to high-dimensional problems and large datasets.

\section{Conclusions and Future Work}
\label{sect:conc}

We have proposed \emph{deep neural rejection} (DNR), \ie, a multi-layer rejection mechanism that, differently from other state-of-the-art rejection approaches against adversarial examples, does not require generating adversarial examples at training time, and it is less computationally demanding. Our approach can be applied to pre-trained network architectures to implement a defense against adversarial attacks at test time. The base classifiers and the combiner used in our DNR approach are trained separately. As future work, it would be interesting to perform an end-to-end training of the proposed classifier similarly to the approaches proposed in~\cite{thulasidasan19-iclr-rejected} and~\cite{geifman19-iclm}. Another research direction may be that of testing our defense against training-time poisoning attacks~\cite{biggio12-icml,jagielski18-sp,biggio15-icml,mei15-aaai,biggio18}.

\vspace{5pt}
\section*{Acknowledgements}
This work was partly supported by the PRIN 2017 project RexLearn, funded by the Italian Ministry of Education, University and Research (grant no. 2017TWNMH2); and by the EU H2020 project ALOHA, under the European Union’s Horizon 2020 research and innovation programme (grant no. 780788).

\balance

\end{document}